\documentclass[lettersize,journal]{IEEEtran}
\usepackage{multirow}
\usepackage[normalem]{ulem}
\useunder{\uline}{\ul}{}
\usepackage{amsmath,amsfonts}
\usepackage{algorithmic}
\usepackage{algorithm}
\usepackage{array}
\usepackage{amssymb}
\usepackage[caption=false,font=normalsize,labelfont=sf,textfont=sf]{subfig}
\usepackage{textcomp}
\usepackage{stfloats}
\usepackage{url}
\usepackage{verbatim}
\usepackage{graphicx}
\usepackage{cite}
\usepackage{booktabs}
\usepackage{caption}
\begin{document}

\title{Text-Audio-Visual-conditioned Diffusion Model for Video Saliency Prediction}

\author{
Li Yu, 
Xuanzhe Sun,
Wei Zhou, \IEEEmembership{Senior Member, ~IEEE}
Moncef Gabbouj, \IEEEmembership{Fellow, ~IEEE}
\thanks{Li Yu is with the School of Computer Science, Nanjing University of Information Science and Technology, Nanjing 210044, China, and also with the Jiangsu Collaborative Innovation Center of Atmospheric Environment and Equipment Technology, Nanjing University of Information Science and Technology, Nanjing 210044, China. (e-mail: li.yu@nuist.edu.cn)}
\thanks{Xuanzhe Sun is with the School of Computer science, Nanjing University of Information Science and Technology, Nanjing 210044, China.}
\thanks{Wei Zhou is with the School of Computer Science and Informatics, Cardiff University, F244AG Cardiff, U.K.}
\thanks{Moncef Gabbouj is with the Faculty of Information Technology and Communication Sciences, Tampere University, 33101 Tampere, Finland.}
}

\markboth{Journal of \LaTeX\ Class Files,~Vol.~14, No.~8, August~2021}%
{Shell \MakeLowercase{\textit{et al.}}: A Sample Article Using IEEEtran.cls for IEEE Journals}

\maketitle

\begin{abstract}
Video saliency prediction is crucial for downstream applications, such as video compression and human-computer interaction. With the flourishing of multimodal learning, researchers started to explore multimodal video saliency prediction, including audio-visual and text-visual approaches. Auditory cues guide the gaze of viewers to sound sources, while textual cues provide semantic guidance for understanding video content. Integrating these complementary cues can improve the accuracy of saliency prediction. Therefore, we attempt to simultaneously analyze visual, auditory, and textual modalities in this paper, and propose TAVDiff, a Text-Audio-Visual-conditioned Diffusion Model for video saliency prediction. TAVDiff treats video saliency prediction as an image generation task conditioned on textual, audio, and visual inputs, and predicts saliency maps through stepwise denoising. To effectively utilize text, a large multimodal model is used to generate textual descriptions for video frames and introduce a saliency-oriented image-text response (SITR) mechanism to generate image-text response maps. It is used as conditional information to guide the model to localize the visual regions that are semantically related to the textual description. Regarding the auditory modality, it is used as another conditional information for directing the model to focus on salient regions indicated by sounds. At the same time, since the diffusion transformer (DiT) directly concatenates the conditional information with the timestep, which may affect the estimation of the noise. To achieve effective conditional guidance, we propose Saliency-DiT, which decouples the conditional information from the timestep. 
Experimental results show that TAVDiff outperforms existing methods, improving 1.03\%, 2.35\%, 2.71\% and 0.33\% on SIM, CC, NSS and AUC-J metrics, respectively.
\end{abstract}

\section{Introduction}
\IEEEPARstart{V}{ideo} Saliency Prediction (VSP) task aims to predict the focus of human visual attention when watching videos. This technique is not only key to understanding the mechanisms of human visual cognition, but also plays an important role in many practical applications. For example, in video coding \cite{wu2024roi, fathima2023neural}, VSP can guide the Region-of-Interest (ROI)-based coding strategy, which can guarantee the user's visual experience while effectively improving compression efficiency and bandwidth utilization. Similarly, in areas such as video surveillance \cite{duan2024saliency} and target detection \cite{an2023sp}, an accurate VSP model can provide guidance on key information. With the exponential growth of video content and the increasing pursuit of high-quality experience by users, the research and application value of VSP tasks are becoming more and more important.

Early research in video saliency prediction primarily focused on utilizing visual cues such as color, texture, motion, and edge information of video frames. The core idea behind these methods was to simulate the reaction mechanisms of the human visual system to low-level visual stimuli. They were typically based on the assumption that areas with high color contrast, complex texture structures, or significant motion are more likely to capture human attention. During this stage, several representative works emerged. For example, Itti et al. \cite{itti2002model} proposed a biologically-inspired visual model-based prediction method that extracts multi-scale feature maps of color, intensity, and orientation, then employs center-surround difference operations to simulate the neuronal competition process and highlight salient regions based on feature contrasts. Hou et al. \cite{hou2007saliency} proposed the spectral residual method, which transforms the image into the frequency domain, calculates the amplitude spectrum, and defines saliency by contrasting the original spectral magnitude with its smoothed version to pinpoint regions with unexpected spectral signatures. With the rise of deep learning technology, subsequent research \cite{zhang2023multi, zhou2023transformer} began to leverage the powerful feature learning capabilities of neural networks to automatically extract higher-dimensional and more complex visual representations, further improving the accuracy of saliency prediction. However, methods relying solely on visual information have fundamental limitations. They often overlook the fact that human attention is actually influenced by multiple sensory modalities. In complex real-world scenes, it is difficult to comprehensively and accurately simulate human gaze mechanisms using only visual information. Therefore, to predict salient regions more comprehensively and accurately, researchers have begun to explore the joint use of multimodal information.

Existing work on multimodal video saliency prediction consists of two main categories: audio-visual methods and text-visual methods. In contrast to purely visual methods, visual-audio methods\cite{tsiami2020stavis, jain2021vinet, xie2024audio, zhu2024mtcam} by integrating audio information, enable the model to capture attention-shifting phenomena that are difficult to detect by relying on vision but are driven by acoustic events (e.g., a sudden loud noise, specific sound sources or vocal dialog). Meanwhile, research has progressed in jointly analyzing visual and textual information for salience prediction\cite{tang2024cardiff}. This approach is based on an intuitive understanding that video-related textual information, such as video descriptions, subtitles, or relevant keywords, often directly or indirectly reflect the attention and interest points of human viewers, which is consistent with human cognitive processes\cite{liu2011examination, holsanova2022cognitive}. Utilizing this intrinsic connection, the high-level semantics provided by the textual information can guide the model to focus attention on regions or objects in the video content that are closely related to the topic. In-depth analysis reveals that the three modalities, visual, audio and textual, naturally contain complementary mechanisms in driving human attention: the visual modality is better at capturing the underlying, physical feature-driven attentional cues; the audio modality can effectively complement the attentional shifts triggered by the acoustic events; and the textual modality focuses on providing the high-level, semantically-driven attentional guidance. Therefore, in order to fully utilize the complementary advantages among different modalities and achieve more comprehensive, accurate, and robust video saliency prediction, exploring models that can effectively integrate visual, audio, and textual information has become an important and promising research direction.

In addition, most of the existing saliency prediction methods, whether visual-only or audio-visual or text-visual, use discriminative models, which directly learn the complex mapping from the high-dimensional input space to the saliency graph, making the task more difficult. In contrast, generative models learn the data distribution of the saliency graph itself. By transforming the saliency prediction task into an image generation task can simplify the learning process and have stronger representation ability. In recent years, the diffusion model\cite{ho2020denoising}, as an emerging generative model, has achieved impressive results in the fields of image generation\cite{li2024distrifusion, yue2024few}, super-resolution reconstruction\cite{yang2024structure}, and so on. The core idea is to transform the data distribution into a simple known distribution (usually a standard Gaussian distribution) by gradually adding Gaussian noise, and then learn the inverse denoising process, thus realizing the generation from noise to target data. This progressive denoising training method is more stable and can generate high-quality samples. In addition, the diffusion model can more naturally consider data from other modalities (e.g., audio, text) as conditional information, which constrains the final generated saliency map to be more semantically consistent with the multimodal input.

Given the limitations of existing saliency prediction methods in fully utilizing multimodal information, and the great ability demonstrated by diffusion models in image generation task, we treat video saliency prediction as an image generation task conditioned on video, audio, and text, and propose TAVDiff, a Text-Audio-Visual-conditioned Diffusion Model for Video Saliency Prediction. In order to fully exploit the rich semantic information embedded in text, we design a novel text semantic guidance method called Saliency-oriented Image-Text Response (SITR) mechanism. SITR captures semantic cues related to saliency regions in textual description and incorporate them into the saliency generation process. In addition, to optimize the way of introducing conditional information and improve the denoising performance, we propose a denoising network called Saliency-DiT. Unlike the Diffusion Transformer (DiT) that directly concatenate the conditional information with the timestep embedding, Saliency-DiT decouples the injection of conditional information from the timestep processing flow. Specifically, the timestep embedding is still fed into Saliengcy-DiT at the beginning. To incorporate conditional information, we newly introduce a cross-attention mechanism in the middle of the self-attention and MLP layers of each Saliency-DiT module. This design avoids direct mixing of conditional information with timestep information on the input side, thus ensuring that the conditional information can  guide the generation process without affecting the processing of timestep. Meanwhile, thanks to the Transformer backbone, Saliency-DiT has a natural advantage over denoising models based on the U-Net architecture in capturing global contexts and long-range dependencies. Extensive experiments show that the proposed TAVDiff achieves significant performance improvements on several datasets, especially in scenarios where textual information has a strong influence on audience attention. To summarize, the main contributions of this paper are as follows:

\begin{itemize}
    \item We propose TAVDiff, the first diffusion model-based tri-modal saliency prediction framework, which comprehensively analyzes and utilizes text, video, and audio modalities, effectively enhancing saliency prediction accuracy.
    \item We design a novel text semantic guidance method, SITR, which mines semantic cues from text to provide more precise semantic guidance for saliency prediction, thereby enhancing the model’s adaptability to different modal data and improving prediction accuracy.
    \item We introduce a new denoising network, Saliency-DiT, which leverages visual, audio, and text features as conditions to guide image generation. This network incorporates these multimodal constraints more naturally via cross-attention, enabling it to adapt more effectively to the multimodal saliency prediction task.
\end{itemize}

\section{Related Works}
\subsection{Audio-visual saliency prediction}
Audio often appears alongside video and can impact human visual fixation, demonstrating its ability to guide or change people's visual focus. When viewers receive specific audio signals, such as environmental sounds or dialogue, their attention is naturally drawn to the visual elements related to the sound.

Early approaches\cite{min2016fixation} attempted to correlate audio events with visual regions, for instance, using canonical correlation analysis (CCA) to link motion and sound. With the development of deep learning, more sophisticated methods emerged. Some focused on localizing sound sources within the visual scene. For example, STAViS \cite{tsiami2020stavis} employed SoundNet \cite{aytar2016soundnet} for audio feature extraction and used bilinear operations for spatial localization, fusing audio event information with spatio-temporal visual features. A dominant trend involves using encoder-decoder architectures, often leveraging 3D convolutions to capture spatio-temporal dynamics directly from video clips. Vinet \cite{jain2021vinet} and CASP \cite{xiong2023casp} adopted UNet-style structures, integrating audio features typically within the decoder stages. CASP specifically emphasized cross-modal feature interaction to leverage audio-visual temporal consistency. Similarly, TSFP \cite{chang2021temporal} utilized a hierarchical feature pyramid network built upon 3D convolutions to aggregate multi-scale spatio-temporal features for improved prediction. Other works like DAVS \cite{zhu2024discrete} also explore deep fusion mechanisms within similar frameworks.

\begin{figure*}[!t]
\centering
\includegraphics[width=7in]{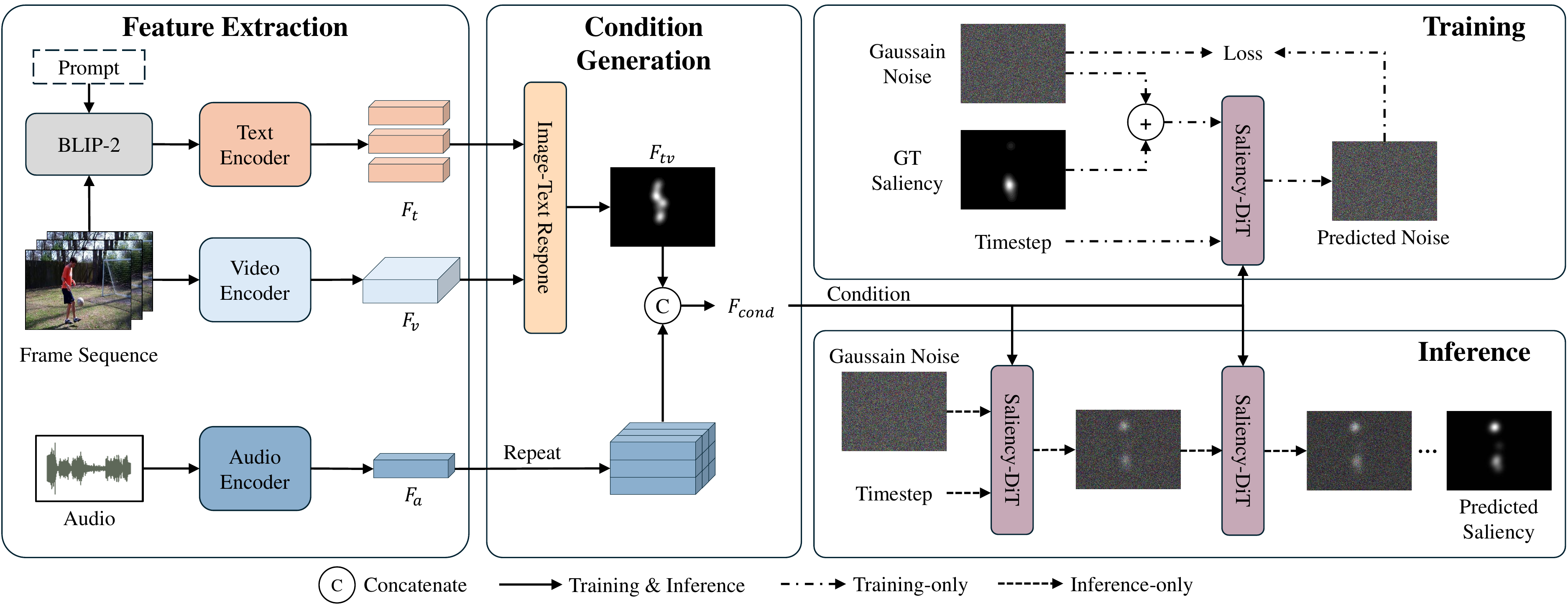}
\caption{The framework of the proposed TAVDiff. The model extracts features from video, audio, and text inputs. A Saliency-oriented Image-Text Response (SITR) mechanism fuses text and visual features. The Saliency-DiT then iteratively generates the saliency map conditioned on audio features $F_a$ and the image-text response $F_{tv}$.}
\label{fig:overview}
\end{figure*}

\subsection{Text-Visual saliency prediction}
Textual modality can provide high-level semantic guidance for visual saliency prediction, and this cross-modal association stems from the intrinsic connection between language and visual attention in human cognitive process. 

Early studies mainly realized semantic guidance through shallow associations between textual keywords and visual features. For example, Ramanishka et al.\cite{ramanishka2017top} proposed the Caption-Guided Visual Saliency approach, which utilizes an image description model to generate textual features, and maps textual semantics to visual space through a cross-modal attentional mechanism to locate saliency regions associated with the description. Similarly, Cornia et al.\cite{cornia2017visual} explored the role of visual saliency in underpinning image descriptions and found that the attention mechanism in a text generation task can inversely enhance the semantic consistency of saliency predictions. Recent work focuses more on exploiting the semantic alignment capability of pre-trained multimodal models. Xue et al.\cite{xue2022ecanet} introduced the text encoder of CLIP model in video saliency prediction, and provided fine-grained semantic constraints for saliency prediction by calculating the semantic similarity between video frames and text descriptions to generate the attentional heatmap. He et al.\cite{he2019human} constructed a multimodal dataset containing human gaze points and text descriptions, verified the strong correlation between text semantics and visual saliency in the spatial distribution, and proposed a hierarchical attention network to realize the text-conditionalized saliency prediction. Tang et al.\cite{tang2024cardiff} proposed CaDiff introduces a textual-visual cross-attention module to highlight textually relevant regions by dynamically weighting visual features with textual descriptions.

To the best of our knowledge, TAVDiff is the first work on video saliency prediction using tri-modal data (audio, video, text). TAVDiff treats saliency prediction as a multimodal conditional generation task that iteratively generates high-quality saliency maps from Gaussian noise by guiding the denoising process through audio, video, and text features.

\section{Methodology}
The workflow of TAVDiff is primarily divided into three stages: feature extraction, condition generation, and iterative denoising generation based on a diffusion model. The overall framework is shown in Fig.\ref{fig:overview}. At feature extraction stage, for the input video frame sequence, audio, and text description, TAVDiff utilizes pre-trained visual, audio, and text encoders to extract their respective modal features. In the conditional feature generation stage, we designed the saliency-oriented Image-Text Response (SITR) mechanism.SITR utilizes a cross-attention mechanism to highlight response features in visual regions associated with text descriptions, thus achieving an accurate mapping from text semantics to visual regions. These saliency-directed image-text response features are used as conditional information for fusion, which is passed on to the subsequent diffusion model. During the training phase, the model learns how to progressively denoise a noisy version of the ground truth saliency map, guided by the extracted multimodal features as conditions. In the inference phase, the model starts from pure Gaussian noise and iteratively denoises it using the Saliency-DiT denoising network to progressively generate the final predicted saliency map. This entire denoising process is conditionally guided by the extracted text, audio, and visual features, ensuring that the generated saliency map accurately captures the salient regions within the video content that are jointly driven by multimodal information. The subsequent subsections will detail the core components of TAVDiff.

\subsection{Image description generation}
Most of the existing multimodal datasets contain only two modalities, visual and audio, and lack textual data to pair with them. In order to evaluate the ability of the proposed TAVDiff in simultaneously utilizing text, audio and video frames for saliency prediction, the pre-trained visual-verbal grand model BLIP-2 \cite{li2023blip} is used to automatically generate textual descriptions corresponding to the video frames. Specifically, by feeding the video data into the BLIP-2 model frame by frame, while giving a prompt: “Please use no more than 20 words to describe this image, focusing on the following two aspects: (1) The foreground objects in the image. (2) What scene, event or theme is the image mainly expressing”, based on which the model is able to generate highly relevant text descriptions based on the input image.

\begin{figure*}[!t]
\centering
\includegraphics[width=6.5 in]{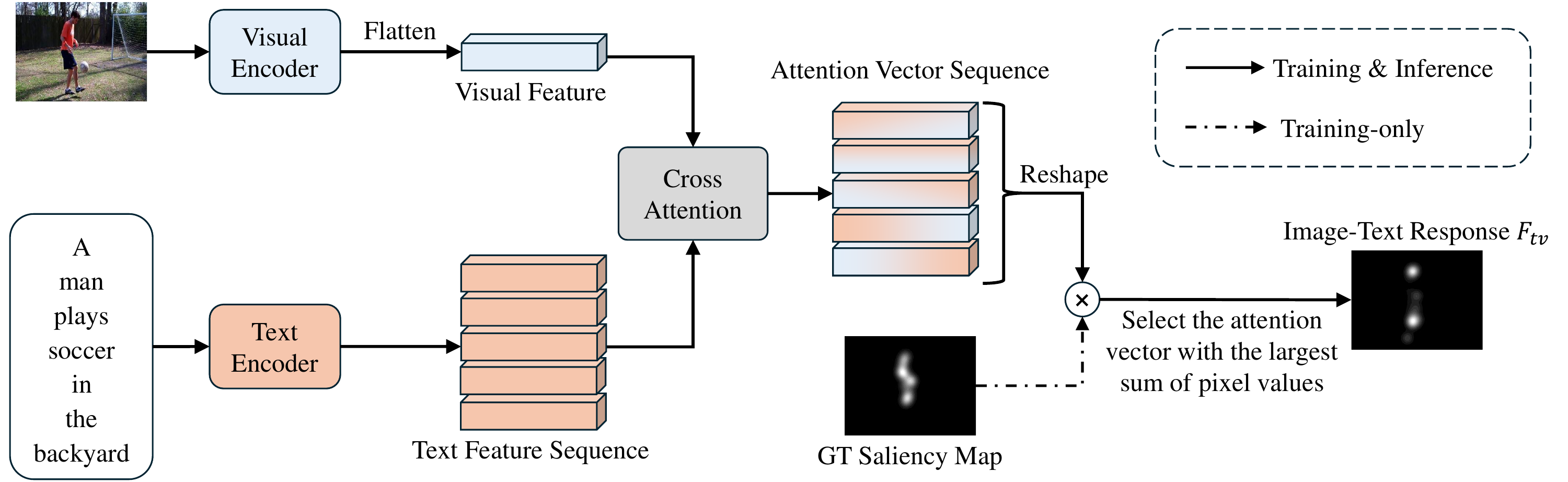}
\caption{The structure of the Saliency-oriented Image-Text Response mechanism (SITR)}
\label{fig:SITR}
\end{figure*}

\subsection{Visual, Audio and Text Backbone}
\textbf{Visual Backbone.}
To effectively extract spatio-temporal features from video, we utilize the S3D network\cite{xie2018rethinking} as the visual backbone. Unlike traditional 2D Convolutional Neural Networks (CNNs), 3D CNNs are capable of simultaneously processing both the spatial and temporal dimensions of video. S3D employs a separable convolution strategy, which not only enhances computational efficiency but also more effectively models spatio-temporal dependencies in videos, thus performing excellently in dynamic video scene processing. Specifically, given the input video frame sequence $V=[v_1, v_2,...,v_{T_v}]$, S3D outputs a spatio-temporal feature $f_v \in R^{t_v \times h_v \times w_v \times c_v}$.

\textbf{Audio Backbone.}
For the audio branch, we retain the original audio data structure as much as possible. Following the approach in \cite{tsiami2020stavis}, we segment the audio waveform according to the duration of the video frames and apply a Hanning window to mitigate edge effects caused by segmentation. For audio feature extraction, we select SoundNet \cite{aytar2016soundnet} as the audio backbone network due to its outstanding performance in learning audio features. Specifically, given the audio waveform $A \in R^{T_a \times 1}$, the audio feature $f_a \in R^{t_a \times c_a}$ is obtained through the SoundNet backbone.

\textbf{Text Encoder.}
For text input, We employ the text encoder from the pre-trained CLIP model \cite{radford2021learning} to extract text features. This encoder utilizes a Transformer architecture, adept at capturing complex textual semantics and contextual information. Crucially, CLIP is trained with a contrastive objective that explicitly aligns text and image representations in a shared embedding space. This inherent cross-modal alignment property ensures the text features are well-suited to interact with visual features. Specifically, let the text sequence be denoted as $T=[t_1, t_2, ..., t_j, ..., t_n]$. CLIP’s text encoder first processes these tokens through an embedding layer to obtain a token embedding sequence $E=[e_1, e_2, ..., e_j, ..., e_n]$, where $e_j \in R^d$. This sequence $E$ is then processed through multi-layer self-attention mechanisms and feed forward neural network within the encoder, ultimately yielding the text feature sequences $F_t=[f_t^1,f_t^2,...,f_t^n]$, where $f_t^j \in R^{c_t}$.

\subsection{Saliency-oriented Image-Text Response Mechanism}
To fully exploit the rich semantic information in text data, we proposes a Saliency-oriented Image-Text Response based text semantic guidance method, as shown in Fig \ref{fig:SITR}. This method effectively captures the semantic correlation between images and text through a multi-head cross-attention mechanism, achieving a deep fusion of textual and visual features. Specifically, the correlation between the text feature sequence $F_t$ and the visual feature $f_v$ is calculated using the multi-head cross-attention mechanism. The attention score between the text feature $f_t^i$ and the visual feature $f_v$ can be expressed as:
\begin{equation}
    score_i=softmax(\frac{Q_t^iK_v}{\sqrt{d_k}})V_v
    \label{eqt:SITR}
\end{equation}
where $Q_t^i$ represents the text query vector obtained by applying an affine transformation to the text feature $f_t^i$. $K_v$ and $V_v$ represent the visual key vectors and visual value vectors, respectively, obtained by applying two other affine transformations to the visual features, $i \in \{1,2,...,n\}$, $d_k$ represents the feature dimension of $K_v$, and $softmax$ is the normalization function. Based on Equation \ref{eqt:SITR}, the multi-head cross-attention mechanism generates a series of attention score maps, each reflecting the spatial semantic correspondence between a specific text feature and the visual features.  During training, each of these attention maps is element-wise multiplied with the ground truth saliency map. This multiplication emphasizes the regions of the image that are both semantically aligned with the corresponding text feature and also considered salient in the ground truth.  From these multiplied maps, the one with the highest sum of pixel values is selected as the most representative Saliency-oriented Image-Text Response map ($f_{tv}$). This selection strategy prioritizes the text feature that best highlights the salient regions within the image according to the ground truth. During inference, without access to the ground truth, the attention map with the highest sum of pixel values is directly selected as $f_{tv}$, representing the strongest semantic correlation between the text and image content. 

\begin{figure*}[!t]
\centering
\includegraphics[width=7 in]{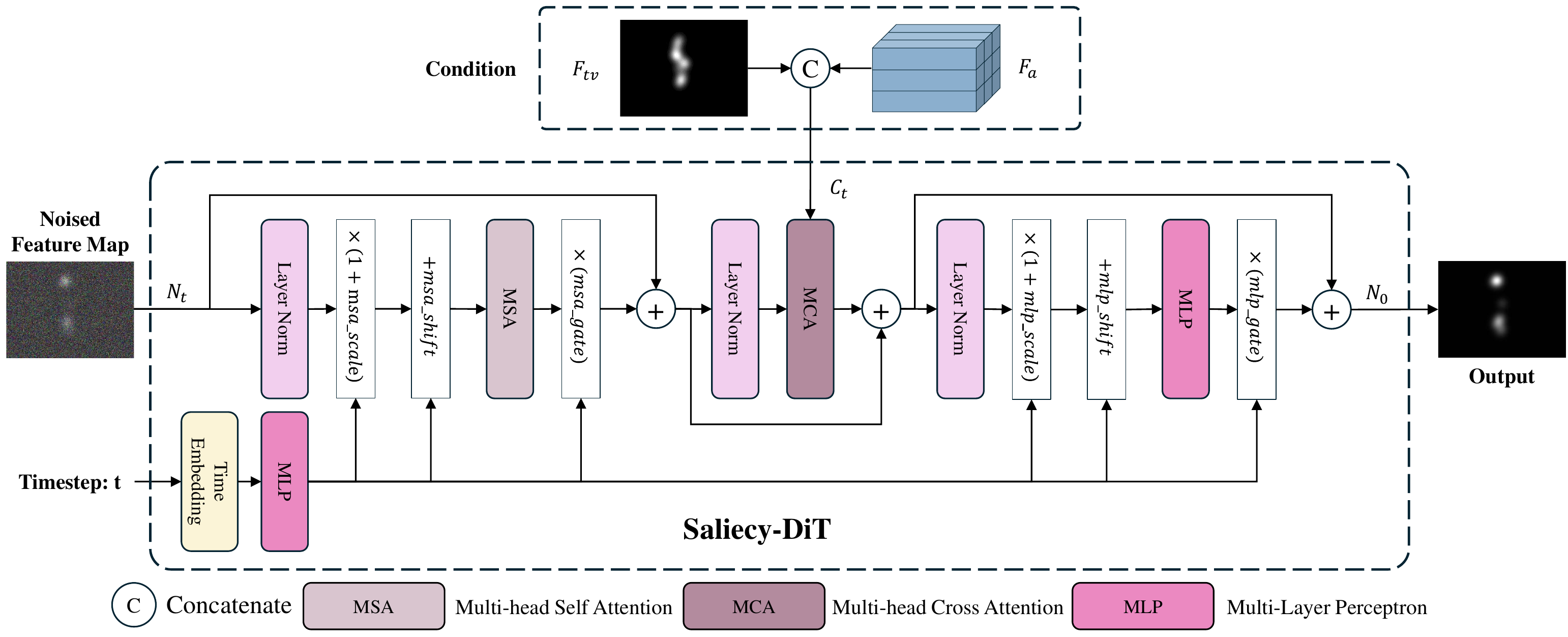}
\caption{The structure of the denoising network Saliency-DiT}
\label{fig:Saliency-DiT}
\end{figure*}

\subsection{Denoising Network Saliency-DiT}
In order to learn the latent distribution of the saliency map, We designed a conditional denoising network $g_\psi$ called Saliency-DiT. Saliency-DiT mainly consists of three parts, namely, the multi-head self-attention phase, the conditional information fusion phase, and the linear projection phase, and its structure is shown in Fig.\ref{fig:Saliency-DiT}.

Saliency-DiT takes three inputs: the diffusion timestep $t$, the noised feature map $N_t$, and the conditional feature $C_t$. The network aims to progressively remove noise from $N_t$ guided by the $C_t$, ultimately producing the fully denoised saliency map $N_0$:
\begin{equation}
    N_0 = g_\psi (N_t, C_t, t)
\end{equation}
Specifically, Saliency-DiT transforms the diffusion timestep $t$ into a high-dimensional representation via a time embedding layer, followed by a multi-layer perceptron (MLP) for non-linear mapping. This generates six adaptive modulation parameters: multi-head self-attention scale $(msa\_scale)$, shift $(msa\_shift)$, and gate $(msa\_gate)$ as well as MLP scale $(mlp\_scale)$, shift$(mlp\_shift)$, and gate $(mlp\_gate)$. These parameters modulate the network based on the diffusion timestep $t$ and are used to dynamically adjust and control the behavior of the subsequent self-attention and MLP modules.  In detail, $msa\_scale$ and $msa\_shift$ are used for an affine transformation of the input to the self-attention module, $msa\_gate$ gates the output of the self-attention module. Similarly, $mlp\_scale$, $mlp\_shift$ and $mlp\_gate$ modulate the MLP module. This modulation allows the network to adapt its denoising strategy based on the different diffusion timestep $t$, leading to more effective denoising.

Concurrently, the noisy image $N_t$ is first processed by Layer Normalization (LayerNorm) to stabilize training and accelerate convergence. Then, a parameterized affine transformation is applied to dynamically scale and shift the input to the self-attention module, adapting it to the noise at the current diffusion timestep $t$:
\begin{equation}
    N_t^1=(1+msa\_scale) \odot LayerNorm(N_t) + msa\_shift.
\end{equation}
where $\odot$ denotes element-wise multiplication. Next, a Multi-Head Self-Attention (MSA) module captures long-range dependencies and contextual information within the image. The output of the self-attention module is further multiplied element-wise with the gating parameter $msa\_gate$:
\begin{equation}
    N_t^2=msa\_gate \odot MSA(N_t^1)
\end{equation}
By using the gating value $msa\_gate$ for feature selection, the influence of redundant attention is suppressed.

Subsequently, the process enters the conditional information fusion stage. The output feature $N_t^2$ from the previous stage is first processed by Layer Normalization, again for feature standardization.  Then, it interacts with the conditional feature $C_t$ through a Multi-Head Cross-Attention (MCA) module:
\begin{equation}
    N_t^3=LayerNorm(MCA(N_t^2,C_t))
\end{equation}
Cross-attention allows the model to effectively integrate the conditional information into the saliency map denoising process, achieving the goal of conditional guidance for saliency map generation.

Finally, the process enters the multi-layer perceptron stage. The features  $N_t^3$ from the cross-attention module are first normalized by LayerNorm and then modulated using a parameterized approach similar to the self-attention stage:
\begin{equation}
    N_t^4 = (1+mlp\_scale) \odot LayerNorm(N_t^3) + mlp\_shift
\end{equation}
where $\odot$ denotes element-wise multiplication. Next, non-linear feature transformation and extraction are performed by the MLP. The output of the MLP is finally multiplied by the gating parameter $ml\_gate$, further controlling the information output of the MLP module through a gating mechanism:
\begin{equation}
    N_t^5 = mlp\_gate \odot N_t^4
\end{equation}
The final output $N_t^5$ is the feature map $N_{t-1}$ after one round of denoising.

\subsection{Training and Inference Process}

\begin{algorithm}
\caption{Training Process of TAVDiff}
\label{alg:TAVDiff_training}
\begin{algorithmic}[1]
\REQUIRE Video frame sequence $V$, Audio $A$, Text sequence $T$, Ground truth saliency map $S_0$
\REPEAT
    \STATE $f_v = \text{VideoEnc}(V)$
    \STATE $f_a = \text{AudioEnc}(A)$
    \STATE $f_t = \text{TextEnc}(T)$
    \STATE $f_{tv} = \text{SITR}(f_v, f_t)$
    \STATE $t \sim \text{Uniform}(1,1000)$; $\epsilon \sim \text{Normal}(0,1)$
    \STATE $S_t = \sqrt{\bar{\alpha}_t}S_0 + \sqrt{1 - \bar{\alpha}_t}\epsilon$
    \STATE Optimization objective: $\Delta_\theta \| g_\psi(S_t, f_a, f_{tv}) - S_0 \|^2$
\UNTIL{Model convergence}
\end{algorithmic}
\end{algorithm}

The training procedure for the proposed TAVDiff is shown in Algorithm \ref{alg:TAVDiff_training}. During the training phase, the model’s objective is to learn how to progressively remove noise from a noisy ground truth saliency map, guided by the provided conditional information, ultimately producing a fully denoised ground truth saliency map. Specifically, the input to TAVDiff during training consists of four types of data: video frame sequence $V$, audio $A$, text sequence $T$, and ground truth saliency map $S_0$. Firstly,  $V$, $A$ and $T$ are passed through their respective modality-specific encoders to obtain visual features $f_v$, audio features $f_a$, and text feature $f_t$. Subsequently, a cross-modal joint representation $f_{tv}$ is generated using the Saliency-oriented Image-Text Response mechanism. Next, a diffusion timestep t is sampled from a uniform distribution, and Gaussian noise $\epsilon$ is sampled from a standard Normal distribution. The ground truth saliency map $S_0$ is then combined with the $\epsilon$ using the noise mixing formula in step 7, resulting in the noisy saliency map $S_t$. The denoising model $g_\psi$ takes $S_t$ and $t$ as its primary inputs, and $f_a$ and $f_{tv}$ as conditional information, to predict the denoised saliency map.  The training of the denoising model uses the Mean Squared Error (MSE) loss between the predicted result and the ground truth saliency map to drive the update of the model parameters until convergence.

\begin{algorithm}
\caption{Inference Procedure of TAVDiff}
\label{alg:TAVDiff_inference}
\begin{algorithmic}[1]
    \REQUIRE Video frame sequence $\mathbf{V}$, audio $\mathbf{A}$, text sequence $\mathbf{T}$, number of inference steps $S$
    \ENSURE Predicted saliency map $\mathbf{S}_{pred}$
    \STATE $f_v = \text{VideoEnc}(\mathbf{V})$
    \STATE $f_a = \text{AudioEnc}(\mathbf{A})$
    \STATE $f_t = \text{TextEnc}(\mathbf{T})$
    \STATE $f_{tv} = \text{SITR}(f_v, f_t)$
    \STATE $S_t \sim \mathcal{N}(0, 1)$
    \STATE $times = \{S \times \frac{1000}{S}, (S-1) \times \frac{1000}{S}, \dots, 1 \times \frac{1000}{S}, 0\}$
    \FOR{$t_{now}, t_{next}$ \textbf{in} Zip($times[:-1]$, $times[1:]$)}
        \STATE $\mathbf{S}_{pred} = g_\psi(S_t, t_{now}, f_a, f_{tv})$
        \STATE $S_t = \text{DDIM}(S_t, \mathbf{S}_{pred}, t_{now}, t_{next})$
    \ENDFOR
\end{algorithmic}
\end{algorithm}

The inference procedure of TAVDiff is shown in Algorithm \ref{alg:TAVDiff_inference}. Unlike the training phase, the model’s input during inference is a pure Gaussian noise map. The model needs to predict salient regions from the Gaussian noise, guided by the provided conditional information. Specifically, the inputs to TAVDiff include: video frame sequence $V$, audio $A$, text sequence $T$, and the number of inference steps $S$. $S$ represents the number of denoising steps required to recover the final saliency map from a completely Gaussian noise map. Similar to the training process, $V$, $A$ and $T$ are passed through their respective encoders to obtain visual features $f_v$, audio features $f_a$, and text feature $f_t$. The Saliency-oriented Image-Text Response mechanism is also used to fuse $f_t$ and $f_v$ into $f_{tv}$. Subsequently, Gaussian noise $S_t$ is sampled from a standard Normal distribution as the denoising target for the model. And a sequence of equally spaced denoising timestep $times$, is constructed based on $S$. The iterative process consists of two main steps: First, the denoising network $g_\psi$ generates a predicted saliency map $S_{pred}$ based on the given audio features $f_a$, the fused features $f_{tv}$, and the current diffusion timestep $t_{now}$. Then, Denoising Diffusion Implicit Models (DDIM) is used to update $S_t$ until $S$ iterations are completed, resulting in the final prediction.

\subsection{Loss Function}
To train the TAVDiff model and ensure that its predicted saliency maps closely approximate the ground truth saliency maps, we select Mean Squared Error (MSE) loss function as the optimization objective for training. The MSE loss function is widely used in regression tasks and can effectively measure the average degree of difference between predicted values and ground truth values. In this research, the MSE loss function is used to measure pixel-level differences between the model’s predicted denoised saliency maps and the ground truth saliency maps, directly optimizing the quality of the saliency maps generated by the model. The calculation formula for the MSE loss function is as follows:
\begin{equation}
    L_{MSE} = \frac{1}{N} \Sigma_{i=1}^N(P_i-G_i)^2
\end{equation}
where $N$ is the total number of pixels in the saliency map, $P_i$ and $G_i$ represent the i-th pixel value of the predicted saliency map and the ground truth saliency map, respectively. By minimizing the MSE loss, TAVDiff can learn to generate predicted results that are highly consistent with the ground truth saliency maps, achieving accurate saliency prediction.

\section{Experiment}
This section begins with a detailed description of the experimental implementation. Subsequently, TAVDiff is compared and analyzed against state-of-the-art saliency prediction methods on widely used datasets within the field, to validate the effectiveness of the TAVDiff network in the video saliency prediction task. Finally, ablation studies are conducted to verify the role of each component in enhancing prediction performance.

\subsection{Dataset}
To evaluate our proposed tri-modal saliency prediction model TAVDiff, we used the commonly used visual-only dataset DHF1K\cite{wang2018revisiting} and six audio-visual datasets: DIEM\cite{mital2011clustering}, Coutrot1\cite{coutrot2014saliency}, Coutrot2\cite{coutrot2016multimodal}, AVAD\cite{min2016fixation}, ETMD\cite{koutras2015perceptually}, and SumMe\cite{gygli2014creating}.

\textbf{DHF1K}: DHF1K consists of 1000 video sequences, which include 600 training videos, 100 validation videos, and 300 test videos. DHF1K encompasses video sequences of various themes, and we chose to pre-train the visual branch of our model on this dataset.

\textbf{Coutrot1}: Coutrot1 enriches the viewer's experience with 60 videos across four thematic categories: individual and group dynamics, natural vistas, and close-ups of faces, supported by visual attention data from 72 contributors.

\textbf{Coutrot2}: In contrast, Coutrot2 focuses on a more niche setting, capturing the interactions of four individuals in a conference setting, with eye-tracking data from 40 observers.

\textbf{DIEM}: The DIEM dataset is even more diverse, containing 84 video clips divided into 64 training and 17 test sets, covering fields such as commercials, documentaries, sports events, and movie trailers. Each video is accompanied by eye-tracking fixation annotations from approximately 50 viewers in a free-viewing mode.

\textbf{AVAD}: The AVAD dataset is a set of 45 brief video sequences, each lasting between 5 and 10 seconds, including a spectrum of dynamic audio-visual experiences, such as musical performances, sports activities, and journalistic interviews. This dataset is enhanced with eye-tracking insights collected from 16 individuals.

\textbf{ETMD}: The ETMD dataset draws its content from a selection of Hollywood cinematic productions, encapsulating 12 distinct film excerpts. This dataset is meticulously annotated with the visual tracking data of 10 evaluators.

\textbf{SumMe}: Completing the collection, the SumMe dataset offers a varied palette of 25 video vignettes, capturing everyday leisure and adventure activities ranging from sports to culinary arts and travel explorations. The visual engagement of viewers is quantified through eye-tracking data from 10 participants.

\subsection{Experimental Details}
\textbf{Experimental Setup.} TAVDiff was trained using the PyTorch framework on a computing platform configured with four NVIDIA GeForce RTX 4090 GPUs in a distributed manner.  We employed the AdamW optimizer \cite{loshchilov2017decoupled} with an initial learning rate set to 1e-4.  The ReduceLROnPlateau learning rate scheduler was used to gradually decrease the learning rate based on the validation loss, with a decay factor of 0.2.  In all experiments, the total diffusion steps were defined as 1000, and the batch size was set to 16. Model training was completed within 80 epochs. During inference, the iterative denoising steps were set to 4.

\textbf{Data Preprocessing.} To adapt the input data to the TAVDiff model and effectively extract audio-visual information from the video content, necessary data preprocessing operations were performed on the raw datasets.

For video data, we extracted 16 consecutive frames from each video clip as the visual input to the model.  All frames were resized to a resolution of $224 \times 384$ to meet the input requirements of the S3D visual backbone network.

For audio data, to ensure temporal alignment with the video frames, we segmented the audio based on the audio sampling rate and the video frame rate, using the following calculation:
\begin{equation}
    N = \frac{A_{freq}}{V_{fps}}
\end{equation}
where $A_{freq}$ is the audio sampling rate and $V_{fps}$ is the video frame rate. This method is based on the premise that the total playback duration of the audio and video should be consistent. By calculating the ratio $N$ of the audio sampling rate to the video frame rate, the number of audio sample points corresponding to each video frame can be determined.  Using $N$, the complete audio can be segmented into frame-aligned clips, thus achieving audio-video synchronization. Furthermore, to mitigate edge effects arising during segmentation, a Hanning window was applied to each extracted audio segment. This windowing function gradually attenuates the signal to zero at both ends, effectively smoothing abrupt changes at frame boundaries and suppressing spectral leakage caused by the truncation of frequency components.

\subsection{Evaluation Metrics}
To compare with existing work, we selected four widely used evaluation metrics: CC, NSS, AUC-J, and SIM. CC is used to measure the linear correlation between the predicted saliency map and the ground truth saliency map. SIM measures the intersection distribution between the predicted saliency maps and the ground truth saliency maps, assessing the degree to which the two distributions match. AUC-J is used to compare the detected saliency map as a binary classifier with the true saliency map. NSS measures the average normalized saliency at the fixed positions of human eye fixations.

\begin{table*}[htb]
\renewcommand{\arraystretch}{1.1}
\caption{Comparison with state-of-the-art methods on the DIEM, ETMD, and AVAD datasets. With best in bold and second best underlined.}
\label{table:comparison with sota part1}
\centering
\scalebox{1.1}{
\begin{tabular}{l|cccc|cccc|cccc}
\hline
\multirow{2}{*}{Method} & \multicolumn{4}{c|}{DIEM}                                        & \multicolumn{4}{c|}{ETMD}                                        & \multicolumn{4}{c}{AVAD}                                         \\
                        & SIM            & CC             & NSS           & AUC-J          & SIM            & CC             & NSS           & AUC-J          & SIM            & CC             & NSS           & AUC-J          \\ \hline
STAViS(AV)              & 0.482          & 0.580          & 2.26          & 0.884          & 0.425          & 0.569          & 2.94          & 0.931          & 0.457          & 0.608          & 3.18          & 0.919          \\
Vinet(AV)               & 0.498          & 0.632          & 2.53          & 0.899          & 0.406          & 0.571          & 3.08          & 0.928          & 0.491          & 0.674          & 3.77          & 0.927          \\
CASP(AV)                & 0.543          & 0.655          & 2.61          & 0.906          & {\ul 0.478}    & {\ul 0.620}    & {\ul 3.34}    & 0.940          & 0.528          & 0.691          & 3.81          & 0.933          \\
TSFP(AV)                & 0.527          & 0.651          & 2.62          & 0.906          & 0.428          & 0.576          & 3.07          & 0.932          & 0.521          & 0.704          & 3.77          & 0.932          \\
DAVS(AV)                & 0.484          & 0.580          & 2.29          & 0.884          & 0.426          & 0.600          & 2.96          & 0.932          & 0.458          & 0.610          & 3.19          & 0.919          \\
Ours(AV)                & 0.547          & 0.670          & 2.75          & 0.909          & 0.446          & 0.613          & 3.15          & {\ul 0.937}    & {\ul 0.550}    & {\ul 0.729}    & {\ul 4.29}    & {\ul 0.949}    \\ \hline
CaDiff(TV)             & 0.544          & 0.664          & 2.70          & 0.905          & 0.443          & 0.600          & 3.04          & 0.922          & 0.531          & 0.719          & 4.18          & 0.933          \\
Ours(TV)                & {\ul 0.552}    & {\ul 0.679}    & {\ul 2.76}    & {\ul 0.910}    & 0.457          & 0.609          & 3.20          & 0.936          & 0.544          & 0.730          & 4.26          & {\ul 0.949}    \\ \hline
Ours(TA)                & 0.341          & 0.476          & 1.54          & 0.699          & 0.225          & 0.379          & 2.11          & 0.648          & 0.367          & 0.517          & 3.18          & 0.734          \\ \hline
Ours(TAV)               & \textbf{0.576} & \textbf{0.688} & \textbf{2.86} & \textbf{0.921} & \textbf{0.489} & \textbf{0.636} & \textbf{3.36} & \textbf{0.941} & \textbf{0.565} & \textbf{0.747} & \textbf{4.34} & \textbf{0.952} \\ \hline
\end{tabular}
}
\end{table*}

\begin{table*}[htb]
\renewcommand{\arraystretch}{1.1}
\caption{Comparison with state-of-the-art methods on the Coutrot1, Coutrot2, and SumMe datasets. With best in bold and second best underlined.}
\label{table:comparison with sota part2}
\centering
\scalebox{1.1}{
\begin{tabular}{l|cccc|cccc|cccc}
\hline
\multirow{2}{*}{Method} & \multicolumn{4}{c|}{Coutrot1}                                    & \multicolumn{4}{c|}{Coutrot2}                                    & \multicolumn{4}{c}{SumMe}                                        \\
                        & SIM            & CC             & NSS           & AUC-J          & SIM            & CC             & NSS           & AUC-J          & SIM            & CC             & NSS           & AUC-J          \\ \hline
STAViS(AV)              & 0.394          & 0.472          & 2.21          & 0.869          & 0.511          & 0.735          & 5.28          & 0.958          & 0.337          & 0.422          & 2.04          & 0.888          \\
Vinet(AV)               & 0.425          & 0.560          & 2.73          & 0.889          & 0.493          & 0.754          & 5.95          & 0.951          & 0.343          & 0.463          & 2.41          & 0.897          \\
CASP(AV)                & 0.456          & 0.561          & 2.65          & 0.889          & 0.585          & 0.788          & 6.34          & 0.963          & \textbf{0.387} & {\ul 0.499}    & {\ul 2.60}    & {\ul 0.907}    \\
TSFP(AV)                & 0.447          & 0.571          & 2.73          & {\ul 0.895}    & 0.528          & 0.743          & 5.31          & 0.959          & 0.360          & 0.464          & 2.30          & 0.894          \\
DAVS(AV)                & 0.400          & 0.482          & 2.19          & 0.869          & 0.512          & 0.734          & 4.98          & 0.960          & 0.339          & 0.423          & 2.29          & 0.889          \\
Ours(AV)                & {\ul 0.486}    & {\ul 0.607}    & 2.85          & 0.892          & {\ul 0.594}    & 0.798          & 6.52          & 0.963          & 0.367          & 0.500          & 2.51          & 0.904          \\ \hline
CaDiff(TV)              & 0.479          & 0.600          & 2.77          & 0.884          & 0.575          & 0.790          & 6.49          & 0.960          & 0.366          & 0.476          & 2.33          & 0.897          \\
Ours(TV)                & 0.476          & 0.602          & {\ul 2.93}    & 0.893          & 0.584          & {\ul 0.802}    & {\ul 6.54}    & {\ul 0.964}    & 0.372          & 0.483          & 2.44          & 0.902          \\ \hline
Ours(TA)                & 0.367          & 0.463          & 2.16          & 0.674          & 0.469          & 0.613          & 4.62          & 0.782          & 0.227          & 0.286          & 1.43          & 0.795          \\ \hline
Ours(TAV)               & \textbf{0.501} & \textbf{0.618} & \textbf{3.05} & \textbf{0.904} & \textbf{0.613} & \textbf{0.819} & \textbf{6.60} & \textbf{0.967} & {\ul 0.384}    & \textbf{0.509} & \textbf{2.66} & \textbf{0.910} \\ \hline
\end{tabular}
}
\end{table*}

\begin{figure*}[htb]
\centering
\includegraphics[width=7 in]{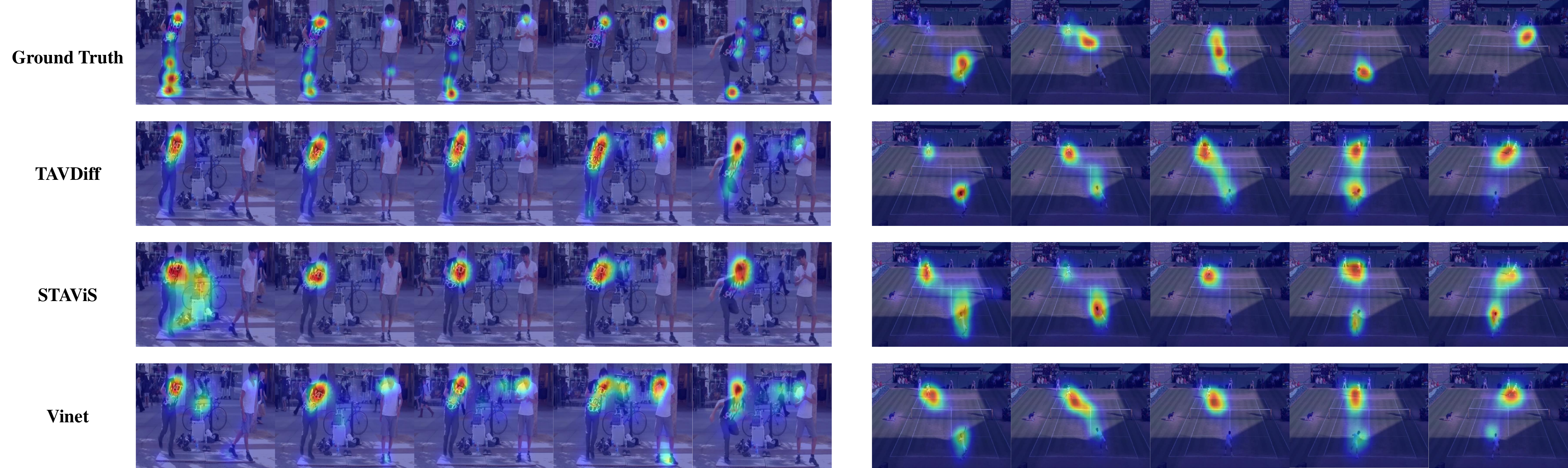}
\caption{Qualitative comparison of the proposed TAVDiff with state-of-the-art methods.}
\label{fig:visualization}
\end{figure*}

\subsection{Performance Comparison}
To validate the effectiveness of the model, the proposed TAVDiff  was comprehensively compared with state-of-the-art methods on six audiovisual saliency datasets. The state-of-the-art methods included in the evaluation are STAViS\cite{tsiami2020stavis}, TSFP\cite{chang2021temporal}, Vinet\cite{jain2021vinet}, CASP\cite{xiong2023casp}, and DAVS\cite{zhu2024discrete}. Considering that most existing methods can only process video and audio data simultaneously, for a fair comparison, we also presents TAVDiff(AV) and its three variants: TAVDiff(TV), TAVDiff(TA), and TAVDiff(TAV), to separately process audio-video pairs, text-video pairs, text-audio pairs, and text-audio-video pairs – representing these four different combinations of multimodal data inputs.

The quantitative comparison results with state-of-the-art methods are shown in Tables \ref{table:comparison with sota part1} and \ref{table:comparison with sota part2}. As can be seen from the tables, TAVDiff(TAV), which simultaneously receives text, audio, and visual inputs, achieved the best results on all six datasets. It achieves an average performance improvement of 2.08\% compared to CASP(AV) and TSFP(AV). Furthermore, even without text input, TAVDiff(AV) achieved results comparable to state-of-the-art methods, further demonstrating the superiority of the proposed method. It is worth noting that TAVDiff(TA), due to the lack of crucial visual input, did not achieve ideal performance. This indicates that saliency prediction is essentially still a task dominated by visual perception, and information from other modalities is difficult to provide sufficient effective cues in a unimodal context. When bimodal combinations are introduced, as long as the combined input includes visual data, the prediction performance can be significantly improved, further confirming the fundamental role of visual information.

In addition, we also presents the qualitative comparison results with state-of-the-art methods, as shown in Fig.\ref{fig:visualization}. From the figure, it can be observed that the saliency maps generated by TAVDiff in various challenging scenarios are closer to the ground truth. For example, in the first column of the left half of Fig.\ref{fig:visualization}, the STAViS and Vinet models mistakenly predicted the bicycle in the background as a salient region, while TAVDiff successfully avoided the interference from the bicycle. The right half of Fig.\ref{fig:visualization} shows a more challenging tennis match scene. In the first column on the right side, the ground truth saliency map indicates that the audience’s attention is mainly focused on the tennis player who is swinging to hit the ball. In contrast, when predicting the tennis scene, the STAViS and Vinet models have a slight deviation in the localization of salient regions, and the concentration of their saliency maps is not as good as the TAVDiff model. In summary, the comparative analysis of different scenes in Fig.\ref{fig:visualization} clearly shows that the TAVDiff model outperforms existing methods in saliency prediction performance in complex and challenging video scenes. Especially in dynamic scenes, TAVDiff can predict salient regions more accurately and stably, which is more consistent with human visual perception.

\subsection{Ablation Study}
\begin{table*}[htb]
\renewcommand{\arraystretch}{1.1}
\caption{Ablation experiment on different denoising networks. With best in bold.}
\label{table:different denoising networks}
\centering
\scalebox{1.1}{
\begin{tabular}{l|cccc|cccc}
\hline
\multirow{2}{*}{Denoising Network} & \multicolumn{4}{c|}{AVAD}                                        & \multicolumn{4}{c}{ETMD}                                         \\
                                   & SIM            & CC             & NSS           & AUC-J          & SIM            & CC             & NSS           & AUC-J          \\ \hline
U-Net                              & 0.542          & 0.728          & 4.20          & 0.935          & 0.554          & 0.669          & 2.73          & 0.910          \\
DiT                                & 0.551          & 0.728          & 4.25          & 0.930          & 0.560          & 0.672          & 2.75          & 0.913          \\
Saliency-DiT                       & \textbf{0.565} & \textbf{0.747} & \textbf{4.34} & \textbf{0.952} & \textbf{0.489} & \textbf{0.636} & \textbf{3.36} & \textbf{0.941} \\ \hline
\end{tabular}
}
\end{table*}

\begin{table*}[htb]
\renewcommand{\arraystretch}{1.1}
\caption{Ablation experiment on Saliency-DiT with and without decoupling of conditional information. With best in bold.}
\label{table:decoupling_ablation}
\centering
\scalebox{1.1}{
\begin{tabular}{c|cccc|cccc}
\hline
\multirow{2}{*}{Saliency-DiT Configuration} & \multicolumn{4}{c|}{AVAD}                                        & \multicolumn{4}{c}{ETMD}                                         \\
                                            & SIM            & CC             & NSS           & AUC-J          & SIM            & CC             & NSS           & AUC-J          \\ \hline
Undecoupled                                 & 0.538          & 0.718          & 4.16          & 0.923          & 0.466          & 0.611          & 3.16          & 0.923          \\
Decoupled                          & \textbf{0.565} & \textbf{0.747} & \textbf{4.34} & \textbf{0.952} & \textbf{0.489} & \textbf{0.636} & \textbf{3.36} & \textbf{0.941} \\ \hline
\end{tabular}
}
\end{table*}

\begin{table*}[htb]
\renewcommand{\arraystretch}{1.1}
\caption{Ablation experiment on different text utilization methods. With best in bold.}
\label{table:SITR_ablation}
\centering
\scalebox{1.1}{
\begin{tabular}{c|cccc|cccc}
\hline
\multirow{2}{*}{Text Utilization Methods} & \multicolumn{4}{c|}{AVAD}                                        & \multicolumn{4}{c}{ETMD}                                         \\
                                          & SIM            & CC             & NSS           & AUC-J          & SIM            & CC             & NSS           & AUC-J          \\ \hline
Concatenate                               & 0.553          & 0.732          & 4.25          & 0.933          & 0.478          & 0.623          & 3.29          & 0.922          \\
Repeat+Add                                & 0.547          & 0.724          & 4.21          & 0.924          & 0.473          & 0.615          & 3.25          & 0.913          \\
Repeat+Average                            & 0.541          & 0.716          & 4.16          & 0.914          & 0.468          & 0.607          & 3.21          & 0.904          \\
SITR                                      & \textbf{0.565} & \textbf{0.747} & \textbf{4.34} & \textbf{0.952} & \textbf{0.489} & \textbf{0.636} & \textbf{3.36} & \textbf{0.941} \\ \hline
\end{tabular}
}
\end{table*}

To validate the effectiveness of various configurations of the proposed TAVDiff, comprehensive ablation experiments were conducted on the AVAD and ETMD datasets. The specific experiments are detailed below:

\textbf{The Effectiveness of Different Denoising Networks.} This experiment aims to verify the impact of different denoising network architectures on the performance of the TAVDiff model. Three denoising networks were compared: UNet [40], DiT [56], and the proposed Saliency-DiT. The experimental results are shown in Table \ref{table:different denoising networks}. The result clearly demonstrates that Saliency-DiT achieves the best performance across all evaluation metrics on both the AVAD and ETMD datasets, significantly outperforming UNet and DiT. Specifically, on the AVAD dataset, the SIM, CC, NSS, and AUC-J performance metrics of Saliency-DiT improved by at least 1.8\% compared to UNet and at least 2.1\% compared to DiT. On the ETMD dataset, Saliency-DiT shows a similar performance improvement trend as on the AVAD dataset, with varying degrees of improvement across all metrics. In summary, the proposed Saliency-DiT, through its advanced architecture and multi-modal conditional information fusion mechanism, can more effectively utilize conditional information for video saliency prediction, resulting in significant improvements across multiple key metrics. This fully demonstrates the superiority and effectiveness of Saliency-DiT in multi-modal video saliency prediction tasks.

\textbf{The Effectiveness of Decoupled Conditional Information in Saliency-DiT.} In order to verify the effectiveness of decoupling the processing of conditional information and timestep in Saliency-DiT, we conducted ablation experiments in both decoupled and undecoupled configurations. In the “decoupled” configuration, the timestep information and the condition information are processed as in Section III, the condition information is integrated into the Saliency-DiT module through the cross-attention mechanism, and the timestep is processed in the original way. In the “undecoupled” configuration, we remove the cross-attention module in Saliency-DiT and directly connect the condition information with the timestep embedding and input it into the network, similar to the processing of standard DiT. The experimental results are shown in Table \ref{table:decoupling_ablation}. As can be seen from the results, the decoupled configuration of Saliency-DiT significantly outperforms the undecoupled configuration in all metrics and datasets. On the ETMD dataset, the decoupled configuration similarly resulted in significant performance gains. These results suggest that directly connecting conditional information to timestep information may interfere with the model's accurate estimation of the noise, thereby degrading denoising performance. In contrast, the decoupled Saliency-DiT avoids interfering with the timestep processing through a cross-attention mechanism while ensuring that the condition information effectively guides the denoising process. This design allows the model to utilize the multimodal conditional information more effectively, thus improving the accuracy of video saliency prediction.

\begin{figure*}[htb]
\centering
\includegraphics[width=7 in]{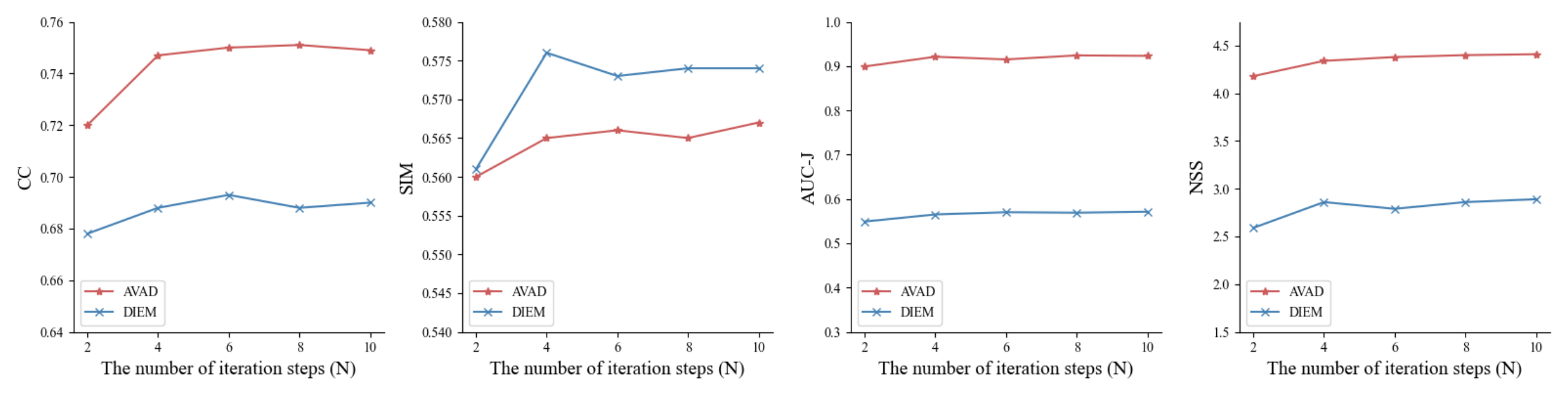}
\caption{Ablation experiments on denoising steps}
\label{fig:plot}
\end{figure*}

\begin{figure}[]
\centering
\includegraphics[width=3 in]{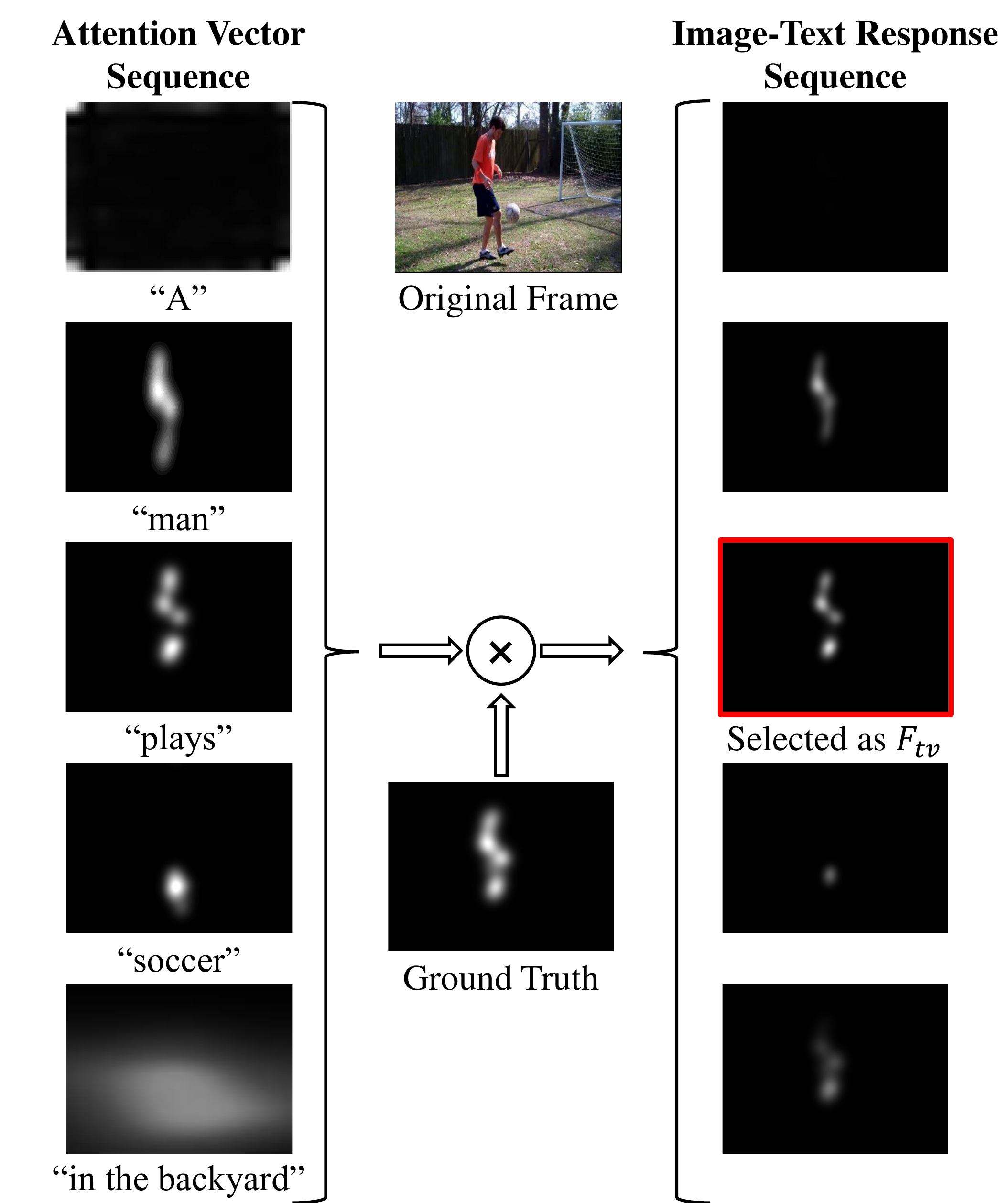}
\caption{Visualization of SITR's workflow.}
\label{fig:ITR_visualization}
\end{figure}

\textbf{The Effectiveness of Different Denoising Iteration.}  Fig.\ref{fig:plot} illustrates the impact of the number of iterative denoising steps on the final performance. Generally, a higher number of iterative denoising steps leads to better overall saliency prediction performance, but with diminishing returns. Initially (N=2 to N=4), increasing the number of iterations significantly improves performance, indicating that a few iterations are crucial for recovering saliency structures from noise. However, as the number of iterations continues to increase (N>4), the performance improvement slows down significantly and even stagnates, suggesting that excessive iteration steps provide limited benefits. Considering the linear increase in computational cost with the number of iterations, a step count of N=4 was chosen to balance performance and efficiency.

\textbf{The Effectiveness of SITR.} To validate the effectiveness of the SITR method, we compared different text utilization strategies. The results are shown in Table \ref{table:SITR_ablation}. SITR consistently outperforms methods such as Concatenate, Repeat+Add, and Repeat+Average on both the AVAD and ETMD datasets, achieving the best performance. This indicates that the SITR method can more effectively capture image-text semantic associations, providing more accurate semantic guidance for saliency prediction. In contrast, simple fusion methods like Concatenate inadequately extract textual semantic information, leading to limited performance gains.  Fig.\ref{fig:ITR_visualization} visualizes the SITR process, demonstrating its ability to accurately locate salient regions corresponding to the text description “a man playing soccer”.  The highlighted regions in the attention map clearly focus on the man and the soccer ball, effectively filtering out irrelevant information.

\section{Conclusion}
Existing video saliency prediction methods often fail to fully utilize the complementary advantages of text, audio and visual modalities. To address this problem, we proposes TAVDiff, a Text-Audio-Visual-conditioned Diffusion Model for video saliency prediction. TAVDiff treats saliency prediction as an image generation task conditioned on text, audio, and video, and generates high-quality saliency maps through stepwise denoising. To effectively utilize textual information, we design the Saliency-oriented Image-Text Response (SITR) mechanism to provide semantic guidance by locating visual regions associated with textual descriptions through cross-modal attention. Meanwhile, to optimize conditional information injection and enhance denoising performance, we propose the Saliency-DiT network. This network fuses multimodal conditions more effectively through cross-attention and avoids interference with timestep processing. Experiments on six datasets show that TAVDiff significantly outperforms current state-of-the-art methods, achieving an average performance improvement of 1.03\%, 2.35\%, 2.71\%, and 0.33\% on the commonly used SIM, CC, NSS, and AUC-J metrics, respectively.

\bibliographystyle{IEEEtran}
\bibliography{reference.bib}
\newpage

\end{document}